\documentclass[10pt,twocolumn,letterpaper]{article}

\usepackage[pagenumbers]{cvpr} 

\usepackage{graphicx}
\usepackage{amsmath}
\usepackage{amssymb}
\usepackage{booktabs}
\usepackage{float}
\usepackage{lipsum}

\usepackage[pagebackref,breaklinks,colorlinks]{hyperref}

\usepackage[capitalize]{cleveref}
\crefname{section}{Sec.}{Secs.}
\Crefname{section}{Section}{Sections}
\Crefname{table}{Table}{Tables}
\crefname{table}{Tab.}{Tabs.}

\def\cvprPaperID{*****} 
\def\confName{CVPR}
\def\confYear{2023}

\begin{document}

\title{Bayesian Decision Making to Localize Visual Queries in 2D}

\author{Syed Asjad\\
Northeastern University\\
{\tt\small asjad.s@northeastern.edu}
\and
Aniket Gupta\\
Northeastern University\\
{\tt\small gupta.anik@northeastern.edu}
\and
Hanumant Singh\\
Northeastern University\\
{\tt\small ha.singh@northeastern.edu}
}
\maketitle

\begin{abstract}
This report describes our approach for the EGO4D 2023 Visual Query 2D Localization Challenge. Our method aims to reduce the number of False Positives (FP) that occur because of high similarity between the visual crop and the proposed bounding boxes from the baseline's Region Proposal Network (RPN). Our method uses a transformer to determine similarity in higher dimensions which is used as our prior belief. The results are then combined together with the similarity in lower dimensions from the Siamese Head, acting as our measurement, to generate a posterior which is then used to determine the final similarity of the visual crop with the proposed bounding box. \\Our code is publicly available at \href{https://github.com/s-m-asjad/EGO4D_VQ2D}{https://github.com/s-m-asjad/EGO4D\_VQ2D}.
\end{abstract}

\section{Introduction}
\label{sec:intro}

Visual Query 2D localization challenge focuses on detecting the last occurrence of the object in a visual crop. The visual crop may or may not consist of objects that have been previously observed during training, which makes the challenge similar to a zero-shot or few-shot detection problem.

The baseline ~\cite{grauman2022ego4d} provides a solution for this challenge by using a Region Proposal Network (RPN) on each frame to propose possible locations of objects. These proposals are then passed through a Siamese Head ~\cite{voigtlaender2020siam} to obtain a score that represents their similarity with the visual crop. The proposed region with the highest similarity is considered to be the location of the object in the frame. This highest similarity score for the object in each frame of the video is then used to create a ~\textit{similarity score signal} as a function of time. Local maxima points are identified on this signal, and the object in the most recent local maxima is tracked in both forward and reverse directions to provide a response track of its final occurrence.

However, a limitation of this approach has been previously identified as the presence of false positives that come as a result of the high similarity of the proposals with the visual crop ~\cite{negativeframes}. This can be due to reasons such as an occluded visual query object, distribution gap between training and evaluation, ambiguous query objects, or other factors such as the background having higher similarity with the object in the proposal box.

While the previous methods for this challenge focus on using one single source for determining similarity, we propose using a fusion of different methods for each frame, where one method complements the other. This approach not only increases the probability of detecting the actual object but also reduces the number of false positives. By employing an additional method, which is less prone to similar failure conditions as the original approach, we are able to achieve more confidence in our obtained similarity scores.

\begin{figure*}
\centering
\includegraphics[width=1\linewidth, height=8cm]{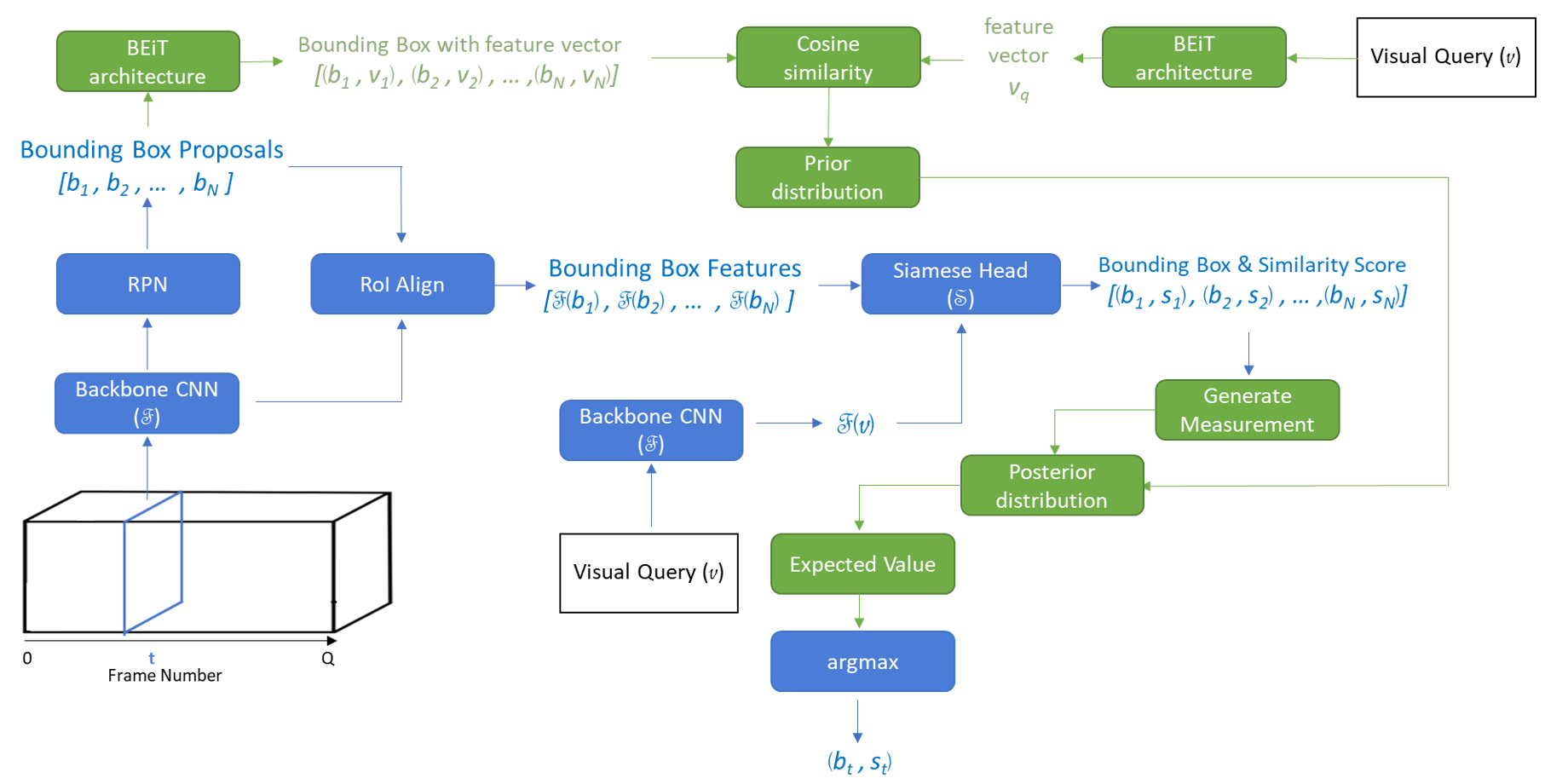}  
\caption{The updated model architecture. Components of the original architecture are shown in blue whilst our improvements are shown in green.}\label{fig:updated_baseline}
\end{figure*}

In our submission, we have proceeded with using the BEiT transformer ~\cite{bao2021beit} to obtain similarity in higher dimensions, which is complemented by the original Siamese Head that performs similarity checking in lower dimensions. The transformer helps extract contextual information from the proposals and visual queries, and is also utilized as a source to form a prior belief regarding the similarity of the two objects. On the other hand, the Siamese Head is used to compute a likelihood distribution and ensure that the object in the proposal box and visual query belong to the same instance. The prior and likelihood are then used to generate a posterior distribution, from which the expected value is considered as the final similarity score for the given proposal box and visual query. Our pipeline is shown in Figure~\ref{fig:updated_baseline}.

\section{Methodology}
\label{sec:methodology}
\subsection{Update to the Baseline}

As mentioned previously, we introduce the BEiT transformer mechanism which encodes the proposal boxes and visual query in higher dimensional space before the similarity score is obtained. The higher dimensional space consist of the BEiT encodings that have been passed through the attention mechanism of the transformer.

The higher dimensional representation of an image by BEiT is a feature vector in $R^+$. We find the cosine similarity of the feature vector for each proposal box with the feature vector of the visual query to generate a prior belief about the presence of the object in the proposal box. \\

\subsection{Generating Distributions}
\subsubsection{Prior}
The shape of the prior distribution was selected to be $Beta(a,b)$ which is one of the most natural distributions for a random variable that is defined between $0$ and $1$ ~\cite{Gupta2011}. The cosine similarity of BEiT feature vectors is treated as the expected value of our prior.

\begin{equation}
    E[x] = \frac{<v_{bbox},v_{VisualQuery}>}{||v_{bbox}||*||v_{VisualQuery}||}
\end{equation}

To generate the shape of the distribution using the expected value, we can rewrite the equation of expected value of the Beta distribution in terms of its hyper-parameters as
\begin{equation}
    E[x] = \frac{a}{a+b}    
\end{equation}

\begin{equation}
    a = b*\frac{E[x]}{1-E[x]}    
\end{equation}

Here, the shape of the distribution can then be generated by deciding the value of the hyper-parameter $b$ which will govern how confident we are with our prior belief.\\

\subsubsection{Measurement}
In order to make our prior act as a conjugate prior, we need to select $Bernoulli$ as the distribution of our measurements. However, $Bern(n,k)$ is a discrete distribution whilst the Siamese head outputs similarity scores which are continuously defined between $0$ and $1$. Furthermore, $n$ being the number of trials and $k$ being the number of successes for those trials would not make sense in our application since we only pass each bounding box through the Siamese head once, and passing it multiple times would not introduce randomization in the result.\\

To resolve the aforementioned challenges, we introduce a way to map our continuous result to this discrete representation by treating our similarity score as a percentage success. As an example, we can treat a similarity score of 0.9 as having success in 9 out of 10 trials, as well as 90 successes out of 100 trials. To determine the number of trials, we introduce a new hyperparameter $w$ which will determine our confidence in the measurement. The similarity score $s$ can be mapped to $n$ and $k$ by

\begin{equation}
    n = w\\
\end{equation}
\begin{equation}
    k = w*s
\end{equation}

\subsection{Determining Final Similarity Score}
We obtain the posterior probability distribution for each proposal box using 
\begin{equation}
    I*p(x|y) \propto I*p(x)p(y|x) = I*Beta(a+k,b+n-k)
\end{equation}

where $I$ is an indicator whose value is $1$ if $E[x]>0.65$ and $s>0.65$, and $0$ otherwise.\\

The similarity score for each proposal box with the visual query can then be estimated by finding the expected value of each posterior
\begin{equation}
    E[p(x|y)] = \frac{a+k}{a+b+n}
\end{equation}

Although we never encountered this situation, but in the case of $E[x]=1$, $a$ from equation$(3)$ would tend to be infinity and become a limitation for modern computers. In such scenario, we proceed further by classifying the situation as one of the following cases.
\begin{itemize}
    \item if one proposal box has $E[x]=1$, we check the Siamese score. If the score is less than $0.65$, we discard the proposal and evaluate the posterior for all other proposal boxes.
    \item if one proposal box has $E[x]=1$ and the Siamese score is more than $0.65$, we discard all other proposals and keep this special proposal box. 
    \item if we have multiple proposal boxes with $E[x]=1$, we use the proposal with the highest Siamese score and discard all other proposals given that the Siamese score was greater than $0.65$.
\end{itemize}

\section{Experiments}
\label{sec:experiments}
We evaluated our approach on the EGO4D evaluation set for the Visual Query 2D task using $b=4.85$ and $w=5$. The two hyper-parameter values were selected by using a sample of clips from the validation set and performing random grid search to tune their values. The size of the sample dataset was kept to 15 clips since it takes 13.5 seconds to process one frame on a V100-sxm2. 

Post-tuning, we optimized the pipeline for inference and the inference speed was now reduced to one frame every 4 seconds. We then evaluated our pipeline, due to time constraints, on a random set of 300 clips from the validation set whilst using the TOMP tracker. The results of our experiment can be viewed in table ~\ref{table:ress}

\begin{table}[ht]
    
\begin{tabular}{|c|c|c|c|c|}
\hline

Method & tAP & stAP & success & recovery\\
\hline
Baseline & 14.12\% & 6.25\% & 43.84\% & 36.71\% \\
\hline
Ours  & ~\textbf{21.68\%} & ~\textbf{9.91\%} & ~\textbf{52.11\%} & ~\textbf{40.98\%} \\    
\hline

\end{tabular}
\caption{Comparison of results using 300 clips from the validation set}
\label{table:ress}
\end{table}

Due to additional time constraints and to be able to evaluate our approach multiple times on the test set before the challenge deadline, we changed the clip's sample rate to 0.5, i.e. we skipped every other frame. Our pipeline was deployed on 20 V100-sxm2s for this evaluation and our results can be viewed in table ~\ref{table:res}. Unfortunately with our current hyperparameters, we were not able to outperform the baseline on the test set using neither the KYS or TOMP tracker.

\begin{table}[ht]
    
\begin{tabular}{|c|c|c|c|c|}
\hline

Method & tAP & stAP & success & recovery\\
\hline
Baseline & 12.57\% & 5.18\% & 41.56\% & 34.03\% \\
\hline
Ours(TOMP)  & 7.39\% & 0.56\% & 30.41\% & 11.55\% \\    
\hline
Ours(KYS)  & 5.30\% & 0.31\% & 24.38\% & 07.64\% \\
\hline

\end{tabular}
\caption{Comparison of test set results between the baseline and our approach}
\label{table:res}
\end{table}

\section{Limitations and Future Work}
The most significant limitation of our approach was the time complexity of the operations and the computational power it requires to execute them. This limitation serves as a major roadblock in using the entire validation dataset to optimize hyper-parameters, and restricts only being able to use a smaller sample of it.

The performance of our approach on the test set can also be attributed to the lack of these optimized hyper-parameters. We have observed that giving a very high weightage to the prior belief (the similarity from BEiT) can cause hallucinations, where for example if the visual query has a metallic object, the pipeline would show high similarity with any random object that has a metallic attribute. On the contrary, giving higher weightage to the Siamese Head brings back the False Positives that occurred due to a lack of contextual information and other previously mentioned reasons.

Finally, it is also worth noting that our approach achieves a significantly lower stAP for a given tAP as compared to the baseline. We expect this to be caused by the proposed bounding box not perfectly surrounding the object in the frame where the most recent local maximum was found, resulting in the accumulation of errors during tracking. The proposed solution for this is a possible modification of the RPN or a run with better hyper-parameters that ensure better matching.

\label{sec:conclusion}

{\small
\bibliographystyle{ieee_fullname}
\bibliography{episodic_memory}
}
\nocite{*}

\end{document}